\documentclass[preprint]{article}
\usepackage{amssymb}

\usepackage[final]{corl_2025} 
\usepackage{xspace}
\usepackage{graphicx}
\usepackage{todonotes}
\usepackage{amsmath}
\usepackage{booktabs}
\usepackage{multirow}
\usepackage{subcaption}
\usepackage{changepage}

\newcommand{\Modelname}{\textsc{BEVCalib}\xspace}
\newcommand{\Datasetname}{\textsc{CalibDB}\xspace}

\newcommand{\ie}{\textit{i.e.,}\xspace}
\newcommand{\eg}{\textit{e.g.,}\xspace}
\newcommand{\etal}{\textit{et al.}\xspace}
\newcommand{\secref}[1]{\S\ref{#1}}
\newcommand{\figref}[1]{Figure~\ref{#1}}
\newcommand{\tabref}[1]{Table~\ref{#1}}

\newcommand{\eqnref}[1]{Equation~\ref{#1}}

\title{\Modelname: LiDAR-Camera Calibration via Geometry-Guided Bird’s-Eye View Representations}

\author{
Weiduo Yuan$^{*1}$, 
Jerry Li$^{*2}$,
Justin Yue$^{2}$,
Divyank Shah$^{2}$,
Konstantinos Karydis$^{2}$,
Hang Qiu$^{2}$
\\
$^{1}$ University of Southern California,
$^{2}$ University of California, Riverside
}

\begin{document}
\maketitle
\makeatletter
\if@conferencefinal
  \def\versiontype{conference}
  \def\thefootnote{*}\footnotetext{Equal contribution. Correspondence to \tt weiduoyu@usc.edu, jli793@ucr.edu}\def\thefootnote{\arabic{footnote}}
\else
  \if@preprinttype
    \def\versiontype{preprint}
    \def\thefootnote{*}\footnotetext{Equal contribution. Correspondence to \tt weiduoyu@usc.edu, jli793@ucr.edu}\def\thefootnote{\arabic{footnote}}
  \else
    \def\versiontype{anonymous}
  \fi
\fi
\makeatother


\begin{abstract}
    Accurate LiDAR-camera calibration is fundamental to fusing multi-modal perception in autonomous driving and robotic systems. Traditional calibration methods require extensive data collection in controlled environments and cannot compensate for the transformation changes during the vehicle/robot movement. In this paper, we propose the first model that uses bird's-eye view (BEV) features to perform LiDAR camera calibration from raw data, termed \Modelname. To achieve this, we extract camera BEV features and LiDAR BEV features separately and fuse them into a shared BEV feature space. To fully utilize the geometric information from the BEV feature, we introduce a novel feature selector to filter the most important features in the transformation decoder, which reduces memory consumption and enables efficient training. Extensive evaluations on KITTI, NuScenes, and our own dataset demonstrate that \Modelname establishes a new state of the art.
    Under various noise conditions, \Modelname outperforms the \textit{best baseline in the literature by an average of (47.08\%, 82.32\%) on KITTI dataset, and (78.17\%, 68.29\%) on NuScenes dataset}, in terms of (translation, rotation), respectively. In the open-source domain, it improves the \textit{best reproducible baseline by one order of magnitude}. %
\def\temp{anonymous}%
\ifx\versiontype\temp All source code and checkpoints will be released.%
\else Our code and demo results are available at \url{https://cisl.ucr.edu/BEVCalib}.

\end{abstract}


\section{Introduction}
Multi-modal sensing has been widely deployed in today's autonomous systems to provide accurate perception while adding redundancy for safety-critical applications. 
Previous work has shown improved reliability and effectiveness of multi-modal perception for navigation in crowded environments~\cite{sathyamoorthy2020densecavoidrealtimenavigationdense} and autonomous driving~\cite{liu2022bevfusion, li2022deepfusionlidarcameradeepfusion} as a result of different sensing modalities complementing each other. One key enabler is the multimodal calibration that ensures the geometric alignment among different modalities. An extrinsic calibration error of a few degrees in rotation or a few cm in translation can compound over a distance (\eg a 20 cm displacement over 5 meters~\cite{9145571}), which can significantly degrade the performance of downstream tasks.

Early works in multimodal calibration relied on targets with unique planar patterns~\cite{zhang2004, yan2023jointcameraintrinsiclidarcamera, 9145571} or specialized rooms~\cite{kitti} as a reference to ensure proper geometry when aligning multiple modalities, primarily image and LiDAR modalities. Although effective, the usage of specialized equipment can make the calibration process tedious and cumbersome. Nevertheless, there is also a demand in modern autonomous systems for continuous calibration in the wild (\eg misoriented/shaken sensors).  Consequently, other works focus on targetless approaches~\cite{schneider2017regnetmultimodalsensorregistration, lv2021lccnetlidarcameraselfcalibration}, \eg relying on the motion of the sensors, using natural features in the environment. The advent of deep learning has further diversified the approaches taken for multimodal calibration. Some calibration methods are hybrid~\cite{koide2023generalsingleshottargetlessautomatic, calibanything}, \ie they use deep learning models to extract features in different modalities and perform traditional optimization to predict the sensor extrinsics. Other methods~\cite{schneider2017regnetmultimodalsensorregistration, lccraft, Herau_2023} are purely data-driven
and are trained and evaluated with popular datasets.
such as KITTI~\cite{kitti}, NuScenes~\cite{nuscenes} and Pandaset~\cite{xiao2021pandasetadvancedsensorsuite}. 
%


Among these learning-based methods, a common pattern is to rely on techniques akin to feature matching between the images and the point clouds. Previous attempts to find these correspondences use feature matching models~\cite{koide2023generalsingleshottargetlessautomatic}, segmentation masks~\cite{calibanything}, or the latent space after encoding images and point clouds as depth images~\cite{schneider2017regnetmultimodalsensorregistration, lv2021lccnetlidarcameraselfcalibration, 9341147, xiao2024calibformertransformerbasedautomaticlidarcamera}. While useful for calibration, establishing correspondences does not explicitly enforce geometric constraints. 
In multi-modal perception works, one appealing method is the bird's-eye-view (BEV) representations~\cite{liu2022bevfusion} that place different modalities in a shared BEV grid. In this BEV grid, LiDAR point clouds are projected or pillarized onto the BEV grid while camera features are also lifted into this space. 
Intrinsically, BEV representations preserve the geometry information, which offers a much stronger space for feature alignment. Such alignment has seen great success in various autonomous driving tasks, including object detection~\cite{li2022bevformer, detr3d, liu2023sparsebevhighperformancesparse3d}, HD-map construction~\cite {li2021hdmapnet, Mask2Map}, place recognition~\cite{BEV-SLAM, luo2023bevplace}, occupancy~\cite{zhang2023occformer, li2024viewformer}, and world model~\cite{zhang2023learning, zhang2024bevworldmultimodalworldmodel}. Therefore, we investigate whether the BEV space is a good candidate for geometric alignment for calibration purposes.

In this work, we propose \Modelname, the first-of-its-kind target-less LiDAR-camera calibration method using BEV representations. This method is motivated by the need to explicitly ensure that geometry is maintained during the calibration process. To that end, \Modelname projects both an input image and a point cloud using an initial guess extrinsic $T_{init}$ into BEV feature space, fuses these BEV features together, and follows a geometry-guided approach to decode $T_{pred}$, the correction needed to arrive at an accurate extrinsic transform. While we train \Modelname on KITTI and NuScenes for fair comparison with existing baselines, we also collect our own dataset (\Datasetname) with heterogeneous extrinsics to evaluate the generalizability. Our evaluation shows that \Modelname establishes a new state-of-the-art performance. Under various noise conditions, \Modelname outperforms the \textit{best baseline in literature} by an average of (47.08\%, 82.32\%) on KITTI dataset, and (78.17\%, 68.29\%) on NuScenes dataset in terms of (translation, rotation) respectively. Compared to open source baselines, \Modelname outperforms the best reproduced results by (92.75\%, 89.22\%) on KITTI dataset, (92.69\%, 93.62\%) on NuScenes dataset, and (60.21\%, 24.99\%) on \Datasetname.  
Qualitative visualizations in the form of camera-LiDAR overlays illustrate a fine-grained projection match as a result of the higher accuracy of \Modelname's predicted extrinsics. 
With strong performance, \Modelname fills a critical gap in the open-source community for LiDAR-camera calibration. 
Our code and demo results are available at \url{https://cisl.ucr.edu/BEVCalib}.



\section{Related Works}


\textbf{Target-based Methods.} Early multimodal calibration methods borrowed from camera calibration techniques using planar targets, \eg checkerboards, fiducial markers, and other specialized patterns, to provide a reference in aligning modalities. Earlier works~\cite{zhang2004} found that LiDAR scans on the planar pattern can be used to register constraints with the estimated pattern on the camera's image plane, thus improving the extrinsic calibration of previous methods. 
Huang~\etal~\cite{9145571} similarly found that using a target of known geometry and dimensions is helpful and developed a solution to fit the LiDAR to camera transform without requiring target edge extraction. Yan~\etal~\cite{yan2023jointcameraintrinsiclidarcamera}  provides a way to jointly calibrate camera intrinsics and LiDAR to camera extrinsics using a special target type with checkerboards and conic sections. Verma \etal~\cite{verma_ITSC2019} proposed using a Variability of Quality (VOQ) metric to score calibration samples, and samples with higher scores are used to reduce user error and possible overfitting to the target. 

\textbf{Target-less Methods.} While specialized targets ensure accuracy in predicting the sensor extrinsic, performing the sensor setup can be cumbersome and tedious. These drawbacks can be alleviated using target-less calibration methods. For example, Ishikawa~\etal~\cite{ishikawa2018lidarcameracalibrationusing} proposed using motion, while Pandey~\etal~\cite{10.5555/2900929.2901018} proposed incorporating probabilistic methods in extrinsic calibration.
Recent years have witnessed the rise of interest in solving calibration using learning-based approaches that leverage natural cues in the target-less setting. Interestingly, the literature follows a divergence of two approaches: combining neural networks with classical methods (\ie hybrid approach), and pure data-driven methods. Hybrid methods~\cite{koide2023generalsingleshottargetlessautomatic, calibanything} use neural networks (\eg SuperGlue~\cite{sarlin2020supergluelearningfeaturematching, kirillov2023segment}) to perform feature extraction before predicting the sensor extrinsic through classical optimization methods. 
Another recent hybrid approach, MDPCalib~\cite{petek2024mdpcalib},  utilizes sensor motion estimates as coarse registration, followed by neural network prediction of 2D-3D correspondence for calibration refinement.
On the other hand, data-driven methods train and evaluate neural networks on datasets such as KITTI~\cite{kitti} and NuScenes~\cite{nuscenes}. Early learning-based methods~\cite{schneider2017regnetmultimodalsensorregistration, lv2021lccnetlidarcameraselfcalibration} encoded the image and LiDAR point cloud before treating the extrinsic prediction as a regression problem. More recently, some works~\cite{Herau_2023, herau2024soacspatiotemporaloverlapawaremultisensor} use neural radiance fields (NeRF~\cite{mildenhall2020nerfrepresentingscenesneural, yang2024unicalunifiedneuralsensor}) as pseudo-targets to ensure explicit geometric alignment between the image and point cloud representations. Furthermore, 3D Gaussian Splatting~\cite{kerbl3Dgaussians}, another volumetric rendering method, is also employed~\cite{herau20243dgscalib} to achieve accurate calibration with more efficient training compared to NeRF.



\textbf{Bird's-eye View Feature.} Bird's-eye view feature space has been used~\cite{li2022bevsurvey, visionbevsurvey} to structure 3D sensor data of the environment into a 2D feature plane. It provides a framework to efficiently extract features from an individual modality~\cite{li2022bevformer, philion2020lift, second} or multiple modalities~\cite{liu2022bevfusion} based on geometric alignment. In recent works, the BEV feature has been adopted to address a wide range of tasks, such as object detection~\cite{detr3d, liu2023sparsebevhighperformancesparse3d}, HD-map construction~\cite {li2021hdmapnet, Mask2Map}, place recognition~\cite{BEV-SLAM, luo2023bevplace}, occupancy perception~\cite{zhang2023occformer, li2024viewformer}, and world model~\cite{zhang2023learning, zhang2024bevworldmultimodalworldmodel}. These works demonstrate the great potential of BEV features on various tasks. Compared to previous works~\cite{lv2021lccnetlidarcameraselfcalibration, xiao2024calibformertransformerbasedautomaticlidarcamera} that use a mis-calibrated depth image as LiDAR input, the BEV feature offers a more accurate and structured geometric representation. 
The closest to our work is CalibRBEV~\cite{calibrbev}, but it only focuses on cameras. The work encodes detection bounding boxes into a BEV representation and applies cross-attention with image features to predict calibration parameters. However, the usage of bounding boxes offers a strong prior knowledge that over-simplifies the calibration process. In contrast, \Modelname calibrates from raw LiDAR data, which is much more challenging but provides more potential for accuracy and robustness to corner cases.
%
To the best of our knowledge, \Modelname is the first cross-modality LiDAR-camera extrinsic calibration model using BEV features.

\vspace{-3mm}
\section{Methodology}
\vspace{-3mm}

\begin{figure}
    \centering
    \includegraphics[width=0.9\linewidth]{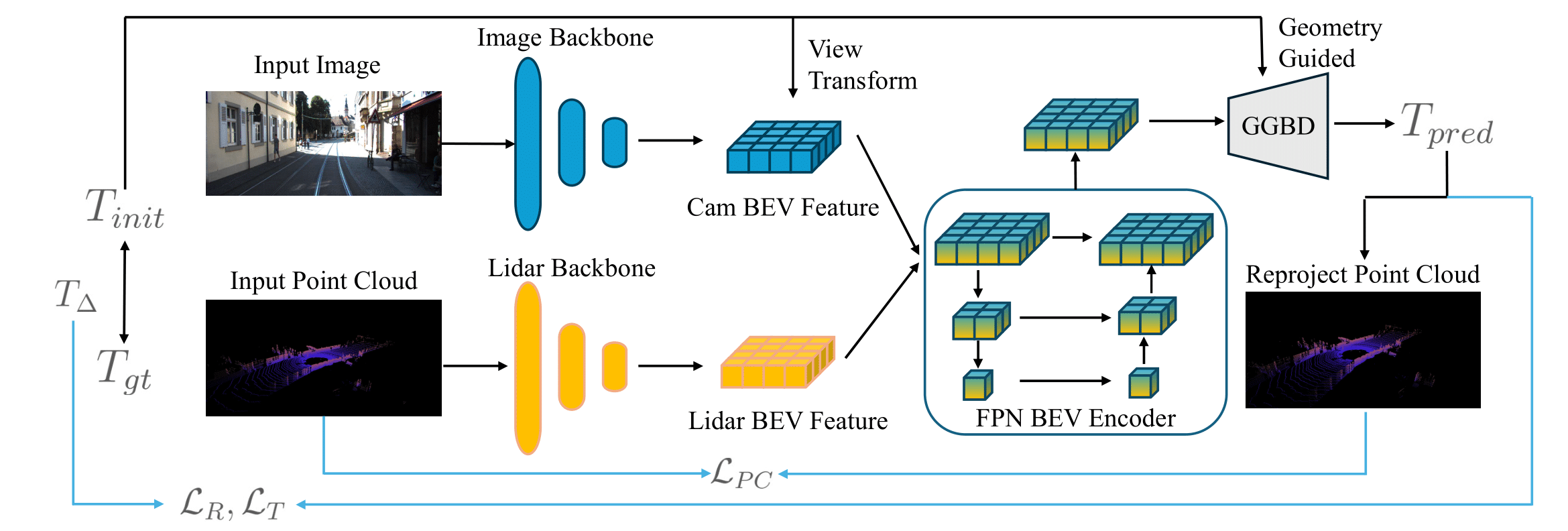}
    \caption{\textbf{Overall architecture of \Modelname.} The overall pipeline of our model consists of BEV feature extraction, FPN BEV Encoder, and geometry-guided BEV decoder (GGBD). For BEV feature extraction (\secref{subsec:bev_feat}), the inputs of the camera and LiDAR are extracted into BEV features through different backbones separately, then fused into a shared BEV feature space. The FPN BEV encoder is used to improve the multi-scale geometric information of the BEV representations. For geometry-guided BEV decoder (\secref{subsec:ggbd}) utilizes a novel feature selector that efficiently decodes calibration parameters from BEV features. $\mathcal{L}_R$, $\mathcal{L}_T$, and $\mathcal{L}_{PC}$ are loss functions introduced at \secref{subsec:loss}.}
    \label{fig:model_overview}
    \vspace{-3mm}
\end{figure}

\subsection{Architecture Overview} 
\Modelname is designed as a target-less LiDAR-camera calibration model that takes a scene consisting of a single image and the full-scene LiDAR data as input and predicts the calibration parameters from LiDAR to camera. 
\figref{fig:model_overview} shows an overall architecture of \Modelname.
It first extracts modality-specific 3D features from camera images and LiDAR using separate backbones~(\secref{subsec:bev_feat}). These features are then projected and fused into a unified BEV representation to capture both semantic and geometric information. To enhance the BEV's spatial capability, we aggregate multi-scale features by a Feature Pyramid Network (FPN) BEV Encoder. Next, we propose a novel Geometry-Guided BEV feature Decoder (GGBD,  \secref{subsec:ggbd}). It first employs a geometry-guided feature selector guided by the coordinates derived from 3D image features, allowing the model to focus on spatially meaningful regions. Finally, it incorporates a refinement module to decode calibration parameters from selected features for efficient and effective training. 
Following the convention of learning-based calibration methods~\cite{calibnet, lv2021lccnetlidarcameraselfcalibration}, \tabref{tab:symbols} summarizes the notations to describe our method.

\begin{table}
\centering
\small
\vspace{-3mm}
\caption{Notation Summary}
\label{tab:symbols}
\begin{tabular}{c|c|l}
Symbol & Dimension & Description \\ 
\hline
\hline
$I$ & $\mathbb{R}^{H \times W \times 3}$ & RGB image captured by camera \\ 
\hline
$P$ & $\mathbb{R}^{N \times 3}$ & Point clouds captured by lidar, where $P_i = [X_i, Y_i, Z_i]$ \\ \hline
$K$ & $\mathbb{R}^{4 \times 4}$ & Intrinsic matrix of camera \\ \hline
$T_{gt}$ & $\mathbb{R}^{4 \times 4}$ & Ground truth transformation from lidar to camera \\ \hline
$T_{\Delta}$ & $\mathbb{R}^{4 \times 4}$ & Random noise input superimposed on $T_{init}$ \\ \hline
$T_{init}$ & $\mathbb{R}^{4 \times 4}$ & Initial guess extrinsic matrix input (including $T_{\Delta}$)  \\ \hline
$T_{pred}$ & $\mathbb{R}^{4 \times 4}$ & Prediction extrinsic matrix as a correction to  $T_{init}$\\ 
\end{tabular}
\vspace{-6mm}
\end{table}


Specifically, the image branch of \Modelname takes image input $I$, and utilizes $T_{init}$ and $K$ to generate a 3D frustum feature $F_{C}^{3D}$ (see more details in \secref{subsec:bev_feat}). Simultaneously, the LiDAR branch encoded LiDAR input $P$ to a voxel feature $F_{L}^{3D}$. These features are then fused into BEV features $F_{\mathcal{B}}$, which is subsequently decoded by GGBD component to get the prediction $T_{pred}$.
%
%
In the training and evaluation process, the initial extrinsic matrix is constructed by superimposing a random noise $T_\Delta$ on top of the groundtruth $T_{gt}$. Hence,  $T_{init}=T_{\Delta}\cdot T_{gt}$ (see more details in \secref{subsec:bev_feat}). Since $T_\Delta$ represents the random noise, a larger $T_\Delta$ means $T_{init}$ will have a larger misalignment and make the problem more challenging. In our setting, we consider various magnitudes of perturbation up to $\{\pm 1.5m, \pm 20^\circ\}$ as the noise range, representing a realistic and challenging calibration scenario.
For evaluation, \Modelname takes $I$, $P$, $K$, and $T_{init}$ as input, output a prediction $T_{pred}$ to compensate for the injected noise. The final LiDAR to camera extrinsic prediction is $\hat{T}_{gt} = T_{pred}^{-1}\cdot T_{init}$. This strategy is useful to control the difficulty of the calibration problem without label leakage.

\subsection{BEV Feature Extraction}
\label{subsec:bev_feat}
BEV feature has an inherent geometric meaning, as each feature in BEV space corresponds to a specific area in the real world. In our setting, we use the LiDAR's coordinate as the world coordinate, which also serves as the BEV coordinate. Inspired by the previous cross-modal approaches~\cite{li2022bevformer}, we adopt a similar paradigm that processes each modality separately and fuses them into a unified BEV feature space. 
Specifically, the LiDAR branch processes the input point cloud $P$ using sparse convolutional backbone to produce a voxel feature $F_{L}^{3D}\in \mathbb{R}^{N_L\times X\times Y \times Z}$, which is then flattened to BEV features $\mathcal{B}_{L}^{2D} \in \mathbb{R}^{(N_L\times Z)\times X\times Y}$, where $X$, $Y$ are the spatial shape of BEV plane, and $Z$ is the number of vertical voxels along the height axis.

The image branch leverages a 2D backbone and an LSS~\cite{philion2020lift} module. The model first extracts the image feature $F_{C}^{2D} \in \mathbb{R}^{f_H\times f_W \times N_C}$ from camera input $I$, where $f_H$, $f_W$ are the shape of image feature. The LSS module defines a discrete depth set for each pixel $(u,v)$, termed as 
$\mathcal{D} = \{d_{min} + \frac{d_{max} - d_{min}}{D - 1} \times i\}_{i=0}^{D - 1}$
, where $D$ is the number of discrete depth bins. For each pixel $(u,v)$, LSS produces $D$ points, accumulating a frustum with $f_H\times f_W \times D$ points in total. The corresponding 3D features are represented as $F_C^{3D} \in \mathbb{R}^{D\times f_H \times f_W \times N_C}$, and the 3D positions in the camera coordinate is defined as $P_C\in \mathbb{R}^{D\times f_H \times f_W \times 3}$. To give the model an initial guess of the position, the frustum coordinates are transformed into world coordinates by $P_C^{W}=[T_{init}^{-1}\cdot \tilde{P}_C]_{1:3}$. Finally, we can get the camera's BEV features $\mathcal{B}_{C}^{2D} \in \mathbb{R}^{N_C \times X \times Y}$ using BEV pooling~\cite{liu2022bevfusion}.


To get a unified BEV representation, we use a $1\times 1$ convolution to fuse features from different modalities, \ie $F_{\mathcal{B}} = \text{Conv1D}([\mathcal{B}_C^{2D}, \mathcal{B}_L^{2D}]) \in \mathbb{R}^{N_\mathcal{B} \times X \times Y}$. We then adopt an FPN BEV Encoder to enhance the multi-scale geometric information of BEV representation.

\subsection{Geometry-Guided BEV Decoder (GGBD)}
\label{subsec:ggbd}
\label{subsec:ggbd}
\begin{figure}
    \centering
    \includegraphics[width=0.85\linewidth]{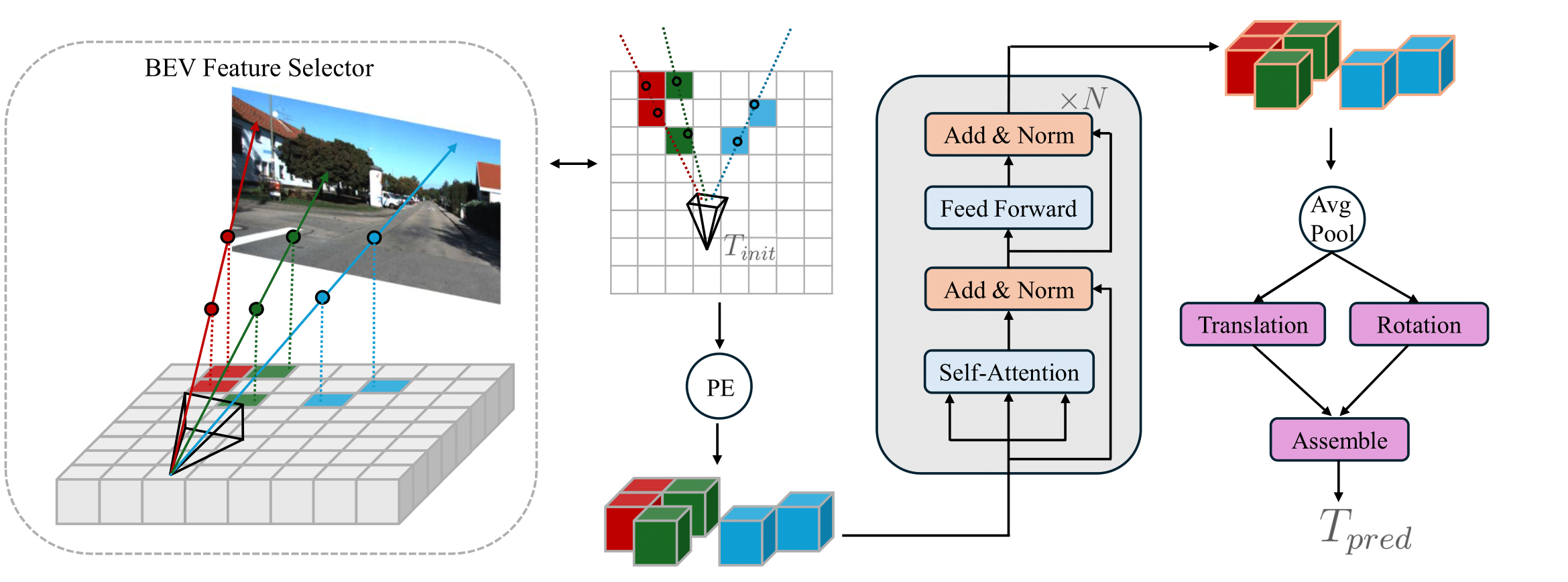}
    \vspace{-1mm}
    \caption{\textbf{Overall Architecture of Geometry-Guided BEV Decoder (GGBD).} The GGBD component contains a feature selector (left) and a refinement module (right). The feature selector calculates the positions of BEV features using \eqnref{eq:proj}. The corresponding positional embeddings (PE) are added to keep the geometry information of the selected feature. After the decoder, the refinement module adds an average-pooling operation to aggregate high-level information, following two separate heads to predict translation and rotation parameters. }
    \label{fig:geom_selector}
    \vspace{-4mm}
\end{figure}

Based on the geometric BEV representation of the scene, we further propose a Geometry-Guided BEV feature Decoder to learn meaningful geometry relationships between the camera and the LiDAR. As illustrated in \figref{fig:geom_selector}, the decoder consists of two stages: a feature selector and a refinement module. 
The BEV feature selector guides the model to focus on the BEV features with meaningful spatial information, while the refinement module aggregates high-level features and helps to predict the final extrinsic parameters.


\textbf{Geometry-Guided BEV Feature Selector.} Specifically for the feature selector, following the image branch of BEV feature extraction, we take the 3D feature positions $P_C^W$ as anchors for cross-modal interaction by projecting them into BEV space. Specifically, for a 3D position $p_c = (x, y, z) \in P_C^W$, its corresponding BEV space coordinate is calculated by
$x_B = \frac{X}{2} + \lfloor \frac{x}{s} \rfloor, \ \ y_B=\frac{Y}{2} + \lfloor\frac{y}{s}\rfloor$
where $s$ is the size of the resolution of BEV's grids. We define the projection operation as $\text{Proj}(p) = (x_B, y_B)$, the set of selected BEV feature positions can be formulated as
\vspace{-1mm}
\begin{align}
    P_\mathcal{B} = \text{Set}(\{\text{Proj}(p) | p\in P_C^W\}) \label{eq:proj}
\end{align}
Since the BEV space is a unified fused space shared by different modalities, such projection positions $(x_B, y_B) \in P_\mathcal{B}$ naturally provide a strong spatial prior for different modalities. This strategy inherently focuses on the overlapping regions between the camera and the LiDAR, acting as an implicit geometric matcher while eliminating redundant features. 
    
\textbf{Refinement Module.} To illustrate the strength and generalizability of our geometric selector, we only use vanilla self-attention~\cite{attention_is_all_you_need} as our refinement module. The whole process of the Geometry-Guided BEV Decoder (GGBD) can be written as
\vspace{-1mm}
\begin{align}
    \text{GGBD}(P_C^W, F_\mathcal{B}) &= \text{Self-Attention}\left(\phi_Q(F_\mathcal{\delta}), \phi_K(F_\mathcal{\delta}),\phi_V(F_\mathcal{\delta}) \right) \\
    F_\delta &=\{F_B[:,x_B, y_B]\mid (x_B, y_B) \in P_\mathcal{B}\}
\end{align}
After GGBD, we apply an average-pooling operation to aggregate the feature. Subsequently, two separate multilayer perceptrons (MLPs) are used to predict translation and rotation, respectively. Finally, the predicted components are assembled into the final prediction $T_{pred}$.

\subsection{Calibration Optimization}
\label{subsec:loss}
\Modelname outputs a translation vector $t \in \mathbb{R}^3$ and a rotation quaternion $r \in \mathbb{R}^4$, the supervision $\hat{r}$ and $\hat{t}$ is derived from 
$\hat{T}_{pred} = T_{init} \cdot T_{gt}^{-1} = \begin{bmatrix}
        \text{Q2M}(\hat{r})&\hat{t}\\
        0&1
    \end{bmatrix}$
, where $\text{Q2M}(\hat{r})$ denotes the rotation matrix converted from quaternion $\hat{r}$.
To effectively optimize the extrinsic calibration, we design a set of loss functions that focus on \textit{rotation-only, translation-only, and joint calibration}.

\textbf{Rotation Loss.} For rotation supervision, we adopt a geodesic loss~\cite{posenet} based on quaternion distance 
$\mathcal{L}_{ang} = 2\text{arctan2}\left(\bigl|\bigl|q_{\Delta}^{(1:3)}\bigr|\bigr|_2, \bigl|q_{\Delta}^{(0)}\bigr|\right)$
, where $q_\Delta = r\cdot \hat{r}^{-1}$ is the relative quaternion between $r$ and $\hat{r}$, $||\cdot||_2$ is $l_2$ norm and $|\cdot|$ is the absolute value. We also utilize a normalization loss to restrict the predicted quaternion $r$ to be a valid rotation \ie 
$\mathcal{L}_{norm} = \left(||r||_2 - 1\right)^2$.
Finally, the rotation loss is 
$\mathcal{L}_R =  \mathcal{L}_{ang} + \lambda_{norm}  \mathcal{L}_{norm}$.

\textbf{Translation Loss.} For translation optimization, we use a Smooth-L1 loss to optimize it. We find that this loss alone is sufficient to optimize translation effectively, therefore, we don't incorporate additional objectives. The translational loss follows
$\mathcal{L}_T = \text{Smooth-L1}\left(t, \hat{t}\right)$.

\textbf{Reprojection Loss.} We use the point cloud reprojection loss introduced by LCCNet~\cite{lv2021lccnetlidarcameraselfcalibration}. Specifically, it can directly supervise the alignment of the transformed point cloud using the predicted translation and rotation jointly, which can be written as $\mathcal{L}_{PC} = \frac{1}{N}\sum_{i=1}^N||T_{gt}^{-1}\cdot T_{pred}^{-1}\cdot T_{init} \cdot \tilde{P}_i - \tilde{P}_i||_2$,
where $N$ is the number of points in the given point cloud $P$. 

\textbf{Total Loss Function.} In summary, the combined loss function is 
$\mathcal{L} = \lambda_{R} \mathcal{L}_R + \lambda_T  \mathcal{L}_T + \lambda_{PC} \mathcal{L}_{PC}$.

\textbf{Implementation Details.}
We utilize sparse convolution \cite{second} as the backbone for LiDAR and adopt Swin-Transformer \cite{swint} combined with LSS \cite{philion2020lift} as the backbone for the camera. For indoor datasets, we constrain the environment range to a 9-meter radius, while for outdoor datasets, we extend the range to 90 meters. 
%
We use a weight vector of ($1.0$, $0.5$, $0.5$) for the ($\mathcal{L}_R$, $\mathcal{L}_T$, $\mathcal{L}_{PC}$) losses, respectively, throughout all training runs. We trained \Modelname using only a single NVIDIA RTX 6000 Ada GPU with a batch size of 16 for 500 epochs on each dataset (\secref{sec:eval}). We applied the AdamW optimizer with a weight decay of $1\text{e}^{-4}$ and an initial learning rate of $5\text{e}^{-5}$, which is decayed by a factor of $0.5$ using a StepLR scheduler. 
\vspace{-1em}
\section{Evaluation}
\label{sec:eval}
\textbf{Datasets.}  
To reproduce and compare with existing approaches, we use two of the most popular benchmarks in the LiDAR-camera calibration literature, KITTI~\cite{kitti} and NuScenes~\cite{nuscenes}. The comparison can contextualize \Modelname with related work. 
%
In the meantime, we also collected our own heterogenous extrinsic dataset \Datasetname.
\Datasetname includes 1244 traces. Each trace contains 12 seconds of continuous frames of image, LiDAR point cloud, and their dynamic extrinsic data recorded at 10 Hz. Our results show that \Modelname generalizes well on \Datasetname while this diversity poses significant challenges for existing calibration methods.

\textbf{Metrics.}
%
We evaluate the translation and rotation error magnitude and break them down along each axis. 
The translation error is calculated as the L1 norm between the prediction and the groundtruth $|t_{gt} - t_{pred}|$. For rotation, we calculate the difference between the rotation matrices of prediction ($R_{pred}$) and groundtruth ($R_{gt}$), \ie $R_{pred}R_{gt}^T$, and extract the Euler angles. 

\textbf{Baselines.}
We compare \Modelname with two sets of baseline results, \textit{original results reported in the publications} and \textit{reproduced results from open-source methods}. 
In the first set,  we include methods in the literature which has a similar evaluation setup (\eg noise range) such that the results can be compared fairly. These baselines include Fu et al.~\cite{fucorl2023}, LCCRAFT~\cite{lccraft}, LCCNet~\cite{lv2021lccnetlidarcameraselfcalibration}, SOAC~\cite{herau2024soacspatiotemporaloverlapawaremultisensor}, 3DGS-Calib~\cite{herau20243dgscalib}, and CalibFormer~\cite{calibformer}. 
In the second set,
we tried our best to exhaust all publicly available and reproducible methods, including CalibAnything~\cite{calibanything}, Koide3~\cite{koide2023generalsingleshottargetlessautomatic}, Regnet~\cite{schneider2017regnetmultimodalsensorregistration}, and CalibNet~\cite{calibnet}. 
We use the official sources of CalibAnything and Koide3, and the officially recommended implementations of Regnet and CalibNet\footnote{~We refer to the recommended unofficial implementations for {CalibNet} (\url{https://github.com/gitouni/CalibNet_pytorch}) and {Regnet} (\url{https://github.com/aaronlws95/regnet}).}.

Notably, LCCNet~\cite{lv2021lccnetlidarcameraselfcalibration} and LCCRAFT~\cite{lccraft} use an iterative refinement approach during inference. Their methods first take a random guess similar to ours, then perform multiple inference passes, with each iteration's output serving as input for the next, progressively refining the calibration parameters. In contrast, our model utilizes a one-stage methodology; therefore, for a fair comparison, we only compare to the single-pass results. Several works are excluded from our evaluation either because they are not reproducible or cannot be compared fairly due to methodology differences. For example,  CalibDepth~\cite{calibdepth} does not report single-pass results. MDPCalib~\cite{petek2024mdpcalib} employs hybrid approaches that needs additional heavy computation.

\begin{table}
  \centering
  \caption{Comparing with Original Results from Literature on KITTI~\cite{kitti} }

  \label{tab:kitti_calib_results}
  \resizebox{\textwidth}{!}{%
  \begin{tabular}{llcccccccc}
    \toprule
    \multirow{2}{*}{\shortstack{Noise\\(Trans. Rot.)}} & \multirow{2}{*}{Method} & \multicolumn{2}{c}{Magnitude$\downarrow$}  & \multicolumn{3}{c}{Translation (cm) $\downarrow$} & \multicolumn{3}{c}{Rotation (°) $\downarrow$} \\
    \cmidrule(lr){3-4} \cmidrule(lr){5-7} \cmidrule(lr){8-10}
    & & $E_t$(cm) & $E_R$(°) & \textbf{X} & \textbf{Y} & \textbf{Z} & \textbf{Roll} & \textbf{Pitch} & \textbf{Yaw} \\
    \midrule
    \multirow{5}{*}{($\pm$1.5m, $\pm$20$^\circ$)}
      & Regnet~\cite{schneider2017regnetmultimodalsensorregistration} & 10.7 & 0.50  & 7 & 7 & 4 & 0.36 & 0.25 & 0.24\\ 
      & Fu et al.~\cite{fucorl2023} & 3.3 & 0.28 & 2.3 $\pm$ 0.5 & 2.0 $\pm$ 0.9 & \textbf{1.2 $\pm$ 0.6} & 0.1 $\pm$ 0.0 & 0.2 $\pm$ 0.0 & 0.2 $\pm$ 0.0 \\ 
      & LCCRAFT~\cite{lccraft} & 37.6 & 1.44 & 31.4 & 12.9 & 16.2 & 1.30 & 0.42 & 0.47 \\ 
      
      & LCCNet~\cite{lv2021lccnetlidarcameraselfcalibration} & 15.0 & 0.94 & 11.8 $\pm$ 14.4 & 5.2 $\pm$ 5.3 & 7.6 $\pm$ 4.1 & 0.2 $\pm$ 0.0 & 0.7 $\pm$ 0.6 & 0.6 $\pm$ 0.5 \\ 
      & \textbf{BEVCalib (Ours)} & \textbf{2.4} & \textbf{0.08} 
                         & \textbf{1.8 $\pm$ 1.6} & \textbf{0.5 $\pm$ 0.5} & 1.5 $\pm$ 2.9
                         & \textbf{0.0 $\pm$ 0.1} & \textbf{0.1 $\pm$ 0.1} & \textbf{0.0 $\pm$ 0.1} \\
    \midrule
    \multirow{5}{*}{($\pm$0.5m, $\pm$5$^\circ$)}
      & CalibAnything~\cite{calibanything} & 9.8 & 0.35 & 5.6 $\pm$ 4.0 & 5.0 $\pm$ 4.4 & 6.3 $\pm$ 6.2 & 0.2 $\pm$ 0.2 & 0.2 $\pm$ 0.1 & 0.2 $\pm$ 0.1 \\ 
      & Koide3~\cite{koide2023generalsingleshottargetlessautomatic} & 21.1 & 0.60 & 6.9 $\pm$ 5.6 & 12.2 $\pm$ 14.9 & 15.7 $\pm$ 9.7 & 0.4 $\pm$ 0.1 & 0.2 $\pm$ 0.1 & 0.4 $\pm$ 0.2 \\ 
      &SOAC~\cite{herau2024soacspatiotemporaloverlapawaremultisensor} & 7.8 & 0.30 & \multicolumn{3}{c}{7.8 $\pm$ 3.5 (xyz together)}  & \multicolumn{3}{c}{0.3 $\pm$ 0.2 (rpy together)}  \\ 
      & 3DGS-Calib~\cite{herau20243dgscalib} & 9.6 & 0.45 & \multicolumn{3}{c}{9.6 $\pm$ 2.1 (xyz together)} & \multicolumn{3}{c}{0.5 $\pm$ 0.2 (rpy together)}\\ 
      & \textbf{BEVCalib (Ours)} & \textbf{2.5} & \textbf{0.06} 
                                         & \textbf{1.8 $\pm$ 1.6} & \textbf{0.3 $\pm$ 0.3} & \textbf{1.7 $\pm$ 3.5}
                        & \textbf{0.0 $\pm$ 0.1} & \textbf{0.1 $\pm$ 0.1} & \textbf{0.0 $\pm$ 0.0} \\
    \midrule
    \multirow{2}{*}{($\pm$0.25m, $\pm$10$^\circ$)} 
      & CalibFormer~\cite{calibformer} & 2.1 & 0.29 & 1.1 & 0.9 & 1.6 & 0.08 & 0.26 & 0.09 \\ 
      
      & \textbf{BEVCalib (Ours)} & \textbf{1.8} & \textbf{0.06} 
                         & 1.5 $\pm$ 1.3 & \textbf{0.3 $\pm$ 0.2} & \textbf{1.0 $\pm$ 2.0}
                         & \textbf{0.0 $\pm$ 0.1} & \textbf{0.1 $\pm$ 0.1} & \textbf{0.0 $\pm$ 0.0} \\
    \midrule
    \multirow{2}{*}{($\pm$0.2m, $\pm$20$^\circ$)}
      & CalibNet~\cite{calibnet} & 8.5 & 0.93 & 4.2 & 1.6 & 7.2 & 0.15 & 0.90 & 0.18 \\ 
      & \textbf{BEVCalib (Ours)} & \textbf{1.8} & \textbf{0.04} 
                         & \textbf{1.4 $\pm$ 1.3} & \textbf{0.2 $\pm$ 0.2} & \textbf{1.0 $\pm$ 2.3}
                         & \textbf{0.0 $\pm$ 0.1} & \textbf{0.0 $\pm$ 0.1} & \textbf{0.0 $\pm$ 0.0} \\
    \bottomrule
  \end{tabular}%
  }
\end{table}
\begin{table}
  \centering
  \caption{Comparing with Original Results from Literature on NuScenes~\cite{nuscenes} }

  \label{tab:nuscenes_calib_results}
  \resizebox{\textwidth}{!}{%
  \begin{tabular}{llcccccccc}
    \toprule
    \multirow{2}{*}{\shortstack{Noise\\(Trans. Rot.)}} & \multirow{2}{*}{Method} & \multicolumn{2}{c}{Magnitude$\downarrow$}  & \multicolumn{3}{c}{Translation (cm) $\downarrow$} & \multicolumn{3}{c}{Rotation (°) $\downarrow$} \\
    \cmidrule(lr){3-4} \cmidrule(lr){5-7} \cmidrule(lr){8-10}
    & & $E_t$(cm) & $E_R$(°) & \textbf{X} & \textbf{Y} & \textbf{Z} & \textbf{Roll} & \textbf{Pitch} & \textbf{Yaw} \\
    \midrule
    \multirow{3}{*}{($\pm$0.5m, $\pm$5$^\circ$)}
      & CalibAnything~\cite{calibanything} & 19.7 & 0.41 &  11.0 $\pm$ 7.4 & 10.0 $\pm$ 5.5 & 13.0 $\pm$ 12.2 & 0.2 $\pm$ 0.1 & 0.3 $\pm$ 0.2 & \textbf{0.2 $\pm$  0.1}\\ 
      & Koide3~\cite{koide2023generalsingleshottargetlessautomatic} & 26.7 & 0.75 & 16.5 $\pm$ 12.1 & 15.6 $\pm$ 13.8  & 14 $\pm$ 11.9 & 0.5 $\pm$ 0.2 & 0.4 $\pm$ 0.3 & 0.4 $\pm$ 0.2 \\ 
       & \textbf{BEVCalib (Ours)} & \textbf{4.3} & \textbf{0.13}
                         & \textbf{1.2 $\pm$ 0.7} & \textbf{6.7 $\pm$ 4.1} & \textbf{2.4 $\pm$ 2.4}
                         & \textbf{0.2 $\pm$ 0.1} & \textbf{0.1 $\pm$ 0.1} & \textbf{0.2 $\pm$ 0.1} \\
    \bottomrule
  \end{tabular}
  }
\end{table}

\textbf{Quantitative Results.} 
\tabref{tab:kitti_calib_results} and \tabref{tab:nuscenes_calib_results} compare \Modelname with the originally reported results from the publications on KITTI and NuScenes datasets. Since each of the existing models was trained and evaluated using different noise settings, we group them into different clusters and evaluate \Modelname under the same noise settings for a fair comparison.
On KITTI dataset, \Modelname has only a few centimeter translation error, outperforming the best baselines by an average of 14.29\% - 78.82\%, and less than $0.1^\circ$ rotation error, outperforming the best baselines by an average of 71.43\% - 95.70\% under various noise conditions. On Nuscenes, \Modelname has a slightly bigger error but still outperforms the best baseline by 78.17\% in translation, 68.29\% in rotation. Notably, although \Modelname is trained under the largest noise ($\pm$1.5m, $\pm$20$^\circ$), it shows extremely robustness when evaluated on smaller noise, overcoming the noise sensitivity that cripples previous methods such as LCCNet~\cite{lv2021lccnetlidarcameraselfcalibration}. 
In addition, \Modelname demonstrates remarkable rotation prediction accuracy for all three angles (roll, pitch, yaw) with error below 0.2$^\circ$, achieving a near-perfect result that outperforms any previous methods.

\tabref{tab:calib_results1} compares \Modelname with the reproducible baselines on KITTI, NuScenes, and \Datasetname. In our exhaustive effort searching for reproducible baselines, we find that the open-source space in this LiDAR-camera calibration domain is rather scarce (very few checkpoints) and underperforming despite the abundant literature. Hence, our open-source effort will significantly improve the performance of publicly available calibration tools. Specifically, \tabref{tab:calib_results1} shows that \Modelname outperforms the best open-source baselines by (92.75\%, 89.22\%) on KITTI dataset and by (92.69\%, 93.62\%) on NuScenes dataset, in terms of (translation, rotation), respectively.
While \Modelname approaches near-zero error on most, if not all, samples, CalibNet and Koide3 struggle with predicting the correct z-component while Regnet and CalibAnything struggle with all components on KITTI and NuScenes datasets. Across the board, when an initial guess is required, a random noise between $[-1.5m, 1.5m]$ and $[-20^\circ, 20^\circ]$ has been applied. 

\begin{figure}
  \centering
  \begin{subfigure}{.5\linewidth}
      \centering
    \includegraphics[width=1.0\textwidth]{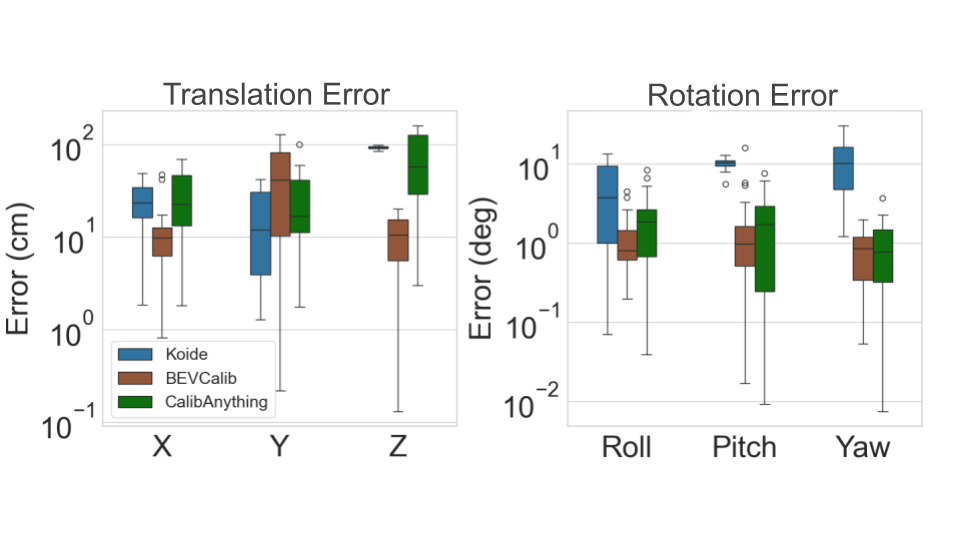}%
      \caption{\Datasetname}
      \label{fig:calibdb-error}
    \end{subfigure}%
    \begin{subfigure}{.5\linewidth}
      \centering
      \includegraphics[width=1.0\textwidth]{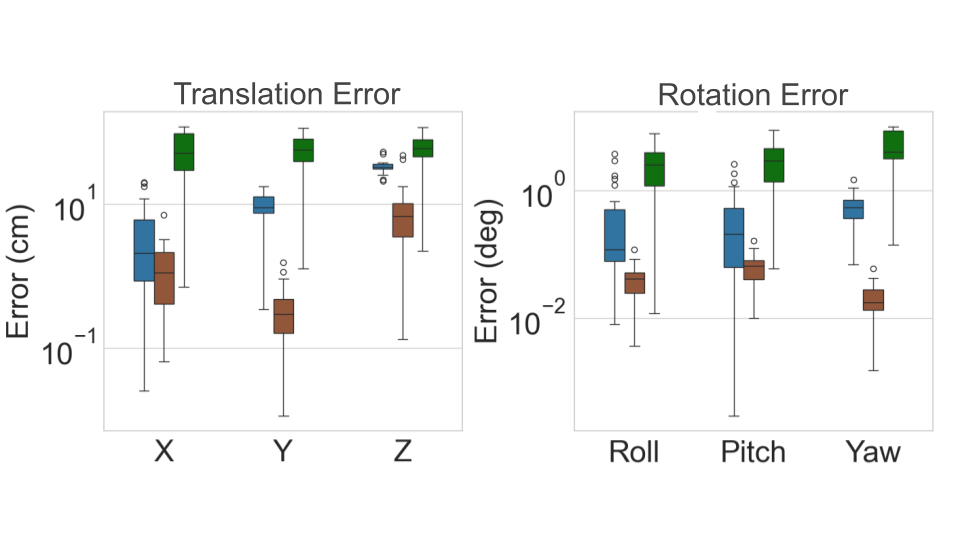}%
      \caption{KITTI}
      \label{fig:kitti-error}
    \end{subfigure}
  \caption{Error Distribution of \Modelname and Other Baselines on \Datasetname and KITTI}
  \label{fig:boxplots}
  \vspace{-6mm}
\end{figure}

On our internal dataset \Datasetname, \Modelname still outperforms the best open-source baselines by (60.21\%, 24.99\%). Compared to KITTI and NuScenes, the error slightly increased for both translation and rotation. This can be attributed to the inherent difficulty of the heterogeneous extrinsics collected in \Datasetname. 
This characteristic is further illustrated in the error distribution shown in \figref{fig:boxplots}. Compared to the error distribution when evaluating on KITTI, there is a larger gap between \Modelname and the baselines evaluated on \Datasetname.

\begin{figure}
    \centering
    \vspace{-6mm}
    \includegraphics[width=1.0\linewidth]{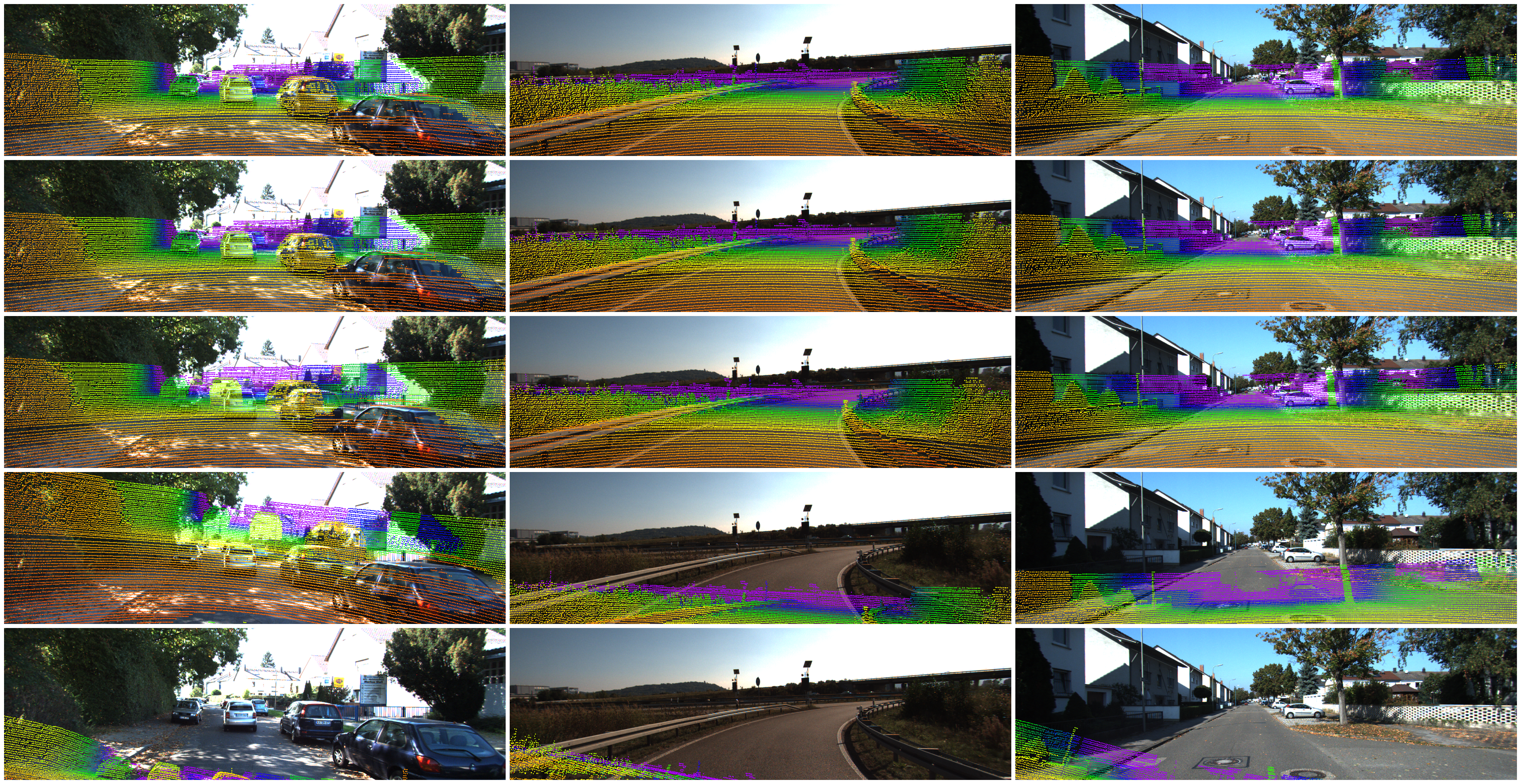}
    \caption{Qualitative results. A comparison of LiDAR-camera overlays from KITTI sequences. From top to bottom: ground-truth, \Modelname, Koide3~\cite{koide2023generalsingleshottargetlessautomatic}, CalibAnything~\cite{calibanything}, Regnet~\cite{schneider2017regnetmultimodalsensorregistration}.}
    \label{fig:overlay-comparison}
\end{figure}

\textbf{Qualitative Results.}
\figref{fig:overlay-comparison} presents a qualitative comparison by overlaying the LiDAR point clouds over the image given each method's predicted extrinsic. Regnet and CalibAnything's overlays are misaligned due to the large error in rotation and translation, so the point cloud is not level with the ground. \Modelname and Koide3 are closer to the ground-truth overlay, but there are objects where Koide3's overlay is slightly misaligned, \eg the misaligned cars in the left column, the traffic sign in the middle column, and the pole and tree in the right column. In contrast, \Modelname's overlays do not show these misalignments. Overall, the overlays reflect the results in \tabref{tab:calib_results1}.

\begin{table}
  \centering
    \caption{Evaluation Results with Reproducible Open-source Baselines
    }
  \label{tab:calib_results1} 

  \resizebox{\textwidth}{!}{%
  \begin{tabular}{llcccccccc}
    \toprule
    \multirow{2}{*}{Dataset} & \multirow{2}{*}{Method}
      & \multicolumn{2}{c}{Magnitude$\downarrow$}
      & \multicolumn{3}{c}{Translation (cm) $\downarrow$}
      & \multicolumn{3}{c}{Rotation (°) $\downarrow$} \\
    \cmidrule(lr){3-4} \cmidrule(lr){5-7} \cmidrule(lr){8-10}
    & & $E_t$(cm) & $E_R$(°) & \textbf{X} & \textbf{Y} & \textbf{Z} & \textbf{Roll} & \textbf{Pitch} & \textbf{Yaw} \\
    \midrule
    \multirow{5}{*}{KITTI~\cite{kitti}}
      & Regnet~\cite{schneider2017regnetmultimodalsensorregistration} & 145.4 & 18.7
      & 101.2 $\pm$ 0.9  &  67.8 $\pm$ 1.2 & 79.4 $\pm$ 0.8 
                         & 16.3 $\pm$ 0.1 & 9.2 $\pm$ 0.0 & 0.8 $\pm$ 0.3 \\ 
      & CalibNet~\cite{calibnet} & 33.1 & 163.7
      & 4.0 $\pm$ 3.8 & 6.5 $\pm$ 4.0 & 32.2 $\pm$ 5.9 & 98.4 $\pm$ 49.9 & 85.9 $\pm$ 2.9 & 98.7 $\pm$ 51.3 \\ 
      & CalibAnything~\cite{calibanything} & 101.1 & 6.8
      & 59.0 $\pm$ 38.0 & 56.0 $\pm$ 30.2 & 60.0 $\pm$ 33.7 & 2.9 $\pm$ 2.1 & 3.2 $\pm$ 2.4 & 5.2 $\pm$ 3.0 \\ 
      & Koide3~\cite{koide2023generalsingleshottargetlessautomatic} & 35.4 & 0.9
      &  4.7 $\pm$ 6.0 & 9.7 $\pm$ 3.9 & 33.7 $\pm$ 6.4 & 0.5 $\pm$ 0.9 & 0.5 $\pm$ 0.6 & 0.6 $\pm$ 0.3 \\ 
      & \textbf{BEVCalib (Ours)} & \textbf{2.4} & \textbf{0.1}
                         & \textbf{1.8 $\pm$ 1.6} & \textbf{0.5 $\pm$ 0.5} & \textbf{1.5 $\pm$ 2.9}
                         & \textbf{0.0 $\pm$ 0.1} & \textbf{0.1 $\pm$ 0.1} & \textbf{0.0 $\pm$ 0.1} \\
    \midrule
    \multirow{5}{*}{NuScenes~\cite{nuscenes}}
      & Regnet~\cite{schneider2017regnetmultimodalsensorregistration} & 196.1 & 93.6
      & 95.2 $\pm$ 0.1 & 72.5 $\pm$ 0.4 & 155.3 $\pm$ 0.1 & 34.4 $\pm$ 0.6 & 71.5 $\pm$ 0.2 & 49.6 $\pm$ 0.4 \\ 
      & CalibNet~\cite{calibnet} & 83.6 & 87.5
      & 5.1 $\pm$ 4.5 & 32.0 $\pm$ 7.4 & 77.1 $\pm$ 5.9 & 87.4 $\pm$ 3.4 & 2.1 $\pm$ 1.9 & 3.0 $\pm$ 2.6 \\ 
      & CalibAnything~\cite{calibanything} & 89.7 & 4.7
      &  58.3 $\pm$ 28.3 & 51.2 $\pm$ 25.4 & 45.0 $\pm$ 30.3 & 2.1 $\pm$ 2.1 & 3.5 $\pm$ 2.8 & 2.4 $\pm$ 2.3 \\ 
      & Koide3~\cite{koide2023generalsingleshottargetlessautomatic} & 82.1 & 149.4
      & 2.2 $\pm$ 1.8 & 32.3 $\pm$ 2.2 & 75.5 $\pm$ 1.8 & 85.2 $\pm$ 38.5 & 88.6 $\pm$ 0.6 & 84.9 $\pm$ 38.8 \\ 
      & \textbf{BEVCalib (Ours)} & \textbf{6.0} & \textbf{0.3}
                         & \textbf{1.3 $\pm$ 1.0} & \textbf{5.4 $\pm$ 4.5} & \textbf{2.3 $\pm$ 2.4}
                         & \textbf{0.2 $\pm$ 0.1} & \textbf{0.2 $\pm$ 0.1} & \textbf{0.2 $\pm$ 0.2} \\
    \midrule
    \multirow{5}{*}{\Datasetname}
      & Regnet~\cite{schneider2017regnetmultimodalsensorregistration} & 216.4 & 24.1
      & 93.0 $\pm$ 27.1& 43.2 $\pm$ 13.9& 190.6 $\pm$ 5.8& 17.6 $\pm$ 5.3& 2.0 $\pm$ 1.3& 16.4 $\pm$ 9.3\\
      & CalibNet~\cite{calibnet} & 95.5 & 180.4 
      &  24.7 $\pm$ 14.1 & 17.4 $\pm$ 13.7& 90.6 $\pm$ 9.2& 72.6 $\pm$ 31.6& 77.3 $\pm$ 4.1& 145.9 $\pm$ 27.5\\ 
      & CalibAnything~\cite{calibanything} & 86.2 & 3.3
      &  31.0 $\pm$ 22.8& 28.4 $\pm$ 25.0& 75.3 $\pm$ 54.2& 2.3 $\pm$ 2.2& 2.1 $\pm$ 2.0 & \textbf{1.0 $\pm$ 0.9} \\
      & Koide3~\cite{koide2023generalsingleshottargetlessautomatic} & 96.8 & 16.5
      &  24.0 $\pm$ 13.1 & \textbf{17.3 $\pm$ 14.4} & 92.2 $\pm$ 4.1 & 5.3 $\pm$ 4.9 & 10.2 $\pm$ 1.7 & 11.9 $\pm$ 9.2 \\
      & \textbf{BEVCalib (Ours)} & \textbf{38.0} & \textbf{2.5}
                         & \textbf{8.4 $\pm$ 11.0} & 36.4 $\pm$ 31.6 & \textbf{6.9 $\pm$ 6.3} 
                         & \textbf{1.2 $\pm$ 1.2} & \textbf{1.7 $\pm$ 2.9} & 1.3 $\pm$ 1.5 \\
    \bottomrule
  \end{tabular}
  }
\end{table}

\textbf{Abalation Study.}
We first conduct an ablation to show the efficacy of the Geometry-Guided BEV feature selector in calibration optimization. The GGBD component (\secref{subsec:ggbd}) consists of a BEV selector and a refinement module. We investigate how different BEV feature selection strategies affect the refinement module. \tabref{tab:ablation_axes} shows that using all BEV features introduces too much redundant information to the model, significantly confusing the model about the cross-modality feature correspondence. 
We also experimented using different attention modules, \eg deformable attention~\cite{DAT}, to capture the relationship between Camera and LiDAR, but the results are less ideal. 

\begin{table}
  \centering
  \small
  \caption{Ablation Results}
  \label{tab:ablation_axes}
  \resizebox{\textwidth}{!}{%
  \begin{tabular}{lcccccc}
    \toprule
    Method
      & \multicolumn{3}{c}{Translation (cm) $\downarrow$}
      & \multicolumn{3}{c}{Rotation (°) $\downarrow$} \\
    \cmidrule(lr){2-4} \cmidrule(lr){5-7}
      & X & Y & Z & Roll & Pitch & Yaw \\
    \midrule
    \Modelname$*$
        & \textbf{8.4 $\pm$ 11.0} & \textbf{36.4 $\pm$ 31.6} & \textbf{6.9 $\pm$ 6.3} & \textbf{1.2 $\pm$ 1.2} & \textbf{1.7 $\pm$ 2.9} & \textbf{1.3 $\pm$ 1.5} \\
    \ $* -$ BEV selector (use all features)
      & 23.1 $\pm$ 19.5 & 37.5 $\pm$ 32.6 & 32.4 $\pm$ 17.6
      & 2.5 $\pm$ 2.4  & 5.0$ \pm$ 3.9  & 2.7 $\pm$ 2.4 \\
    \ $*$ with Deformable Attention
      & 37.0 $\pm$ 30.8 & 37.0 $\pm$ 31.9 & 34.2 $\pm$ 27.3
      & 5.6 $\pm$ 4.8 & 5.3 $\pm$ 4.3 & 5.3 $\pm$ 4.5 \\
    \bottomrule
  \end{tabular}%
  }
\end{table}


\section{Conclusion}

In this paper, we introduce \Modelname, the first LiDAR-camera extrinsic calibration model using BEV features. Geometry-guided BEV decoder can effectively and efficiently capture scene geometry, enhancing calibration accuracy.  Results on KITTI, NuScenes, and our own indoor dataset with dynamic extrinsics illustrate that our approach establishes a new state of the art in learning-based calibration methods. Under various noise conditions, \Modelname outperforms the \textit{best baseline in literature by an average of (47.08\%, 82.32\%) on KITTI dataset, and (78.17\%, 68.29\%) on NuScenes dataset}, in terms of (translation, rotation) respectively. Also, \Modelname improves the \textit{best reproducible baseline by one order of magnitude}, making an important contribution to the scarce open-source space in LiDAR-camera calibration.

\newpage
\bibliography{example}  

\end{document}